\documentclass{article}
\usepackage{spconf,amsmath,epsfig}
\usepackage{amsfonts}
\usepackage{upgreek}
\usepackage{graphicx}
\usepackage{makecell}
\usepackage{array}
\usepackage{units}
\usepackage{multirow}
\usepackage{makecell}
\usepackage{xcolor}
\usepackage{bm}
\usepackage{csvsimple}
\usepackage{pgfplotstable}
\usepackage[hyphens]{url}
\usepackage[hidelinks]{hyperref}
\usepackage[activate={true,nocompatibility},final,tracking=true,kerning=true,spacing=true,factor=1100,stretch=10,shrink=10]{microtype}
\usepackage[square,numbers]{natbib}
\usepackage{caption, setspace}

\newcommand{\specialcell}[2][c]{
  \begin{tabular}[#1]{@{}c@{}}#2\end{tabular}}
\newcommand{\specialcellleft}[2][l]{
  \begin{tabular}[#1]{@{}l@{}}\vspace{-0.02in}#2\end{tabular}}
\graphicspath{{figures/}}
\title{BigEarthNet: A Large-Scale Benchmark Archive for Remote Sensing Image Understanding}

\name{Gencer Sumbul and Beg\"{u}m Demir}
\address{Faculty of Electrical Engineering and Computer Science, Technische Universit\"at Berlin, Germany}

\name{Gencer Sumbul $^1$, Marcela Charfuelan $^2$, Beg\"um Demir $^1$, Volker Markl $^{1,2}$}
\address{$^1$Technische Universit{\"a}t Berlin, $^2$DFKI GmbH} 

\begin{document}
\maketitle
\begin{abstract}
  This paper presents the BigEarthNet that is a new large-scale multi-label Sentinel-2 benchmark archive. 
  The BigEarthNet consists of $590,326$ Sentinel-2 image patches, each of which is a section of i) $120\times120$ pixels 
  for 10m bands; ii) $60\times60$ pixels for 20m bands; and iii) $20\times20$ pixels for 60m bands. 
  Unlike most of the existing archives, each image patch is annotated by multiple land-cover classes 
  (i.e., multi-labels) that are provided from the CORINE Land Cover database of the year 2018 (CLC 2018). 
  The BigEarthNet is significantly larger than the existing archives in remote sensing (RS) and thus is much more convenient to be used as a training source in the context of deep learning. 
  This paper first addresses the limitations 
  of the existing archives and then describes the properties of the BigEarthNet. 
  Experimental results obtained in the framework of RS image scene classification problems show that 
  a shallow Convolutional Neural Network (CNN) architecture trained on the BigEarthNet provides much higher accuracy compared to a state-of-the-art CNN model pre-trained on the ImageNet (which is a very 
  popular large-scale benchmark archive in computer vision). The BigEarthNet opens up promising directions to advance operational 
  RS applications and research in massive Sentinel-2 image archives. 
\end{abstract}

\begin{keywords}
  Sentinel-2 image archive, multi-label image classification, deep neural network, remote sensing
\end{keywords}

\vspace{-0.1in}
\section{Introduction}
\vspace{-0.05in}
\label{sec:intro}
Recent advances in deep learning have attracted great attention in remote sensing (RS) due to the high capability of 
deep networks (e.g., Convolutional Neural Networks (CNN), Recurrent Neural Networks (RNN), Generative 
Adversarial Networks (GAN)) 
to model the high-level semantic content of RS images. 
To train such networks, a very large training set is needed with a high number of annotated images in 
order to learn effective models with several different parameters. 
To the best of our knowledge, publicly available RS image archives contain only a small number of annotated images 
and a large-scale benchmark archive does not yet exist. 
Thus, the lack of a large training set is an important bottleneck that prevents the use of deep learning in RS. 
In order to address this problem, fine-tuning deep networks pre-trained on large-scale computer vision archives 
(e.g., ImageNet) is considered in RS community. However, such an approach has several limitations 
related to the differences on the characteristics of images between computer vision and RS. 
Additionally, in the existing archives, RS images are annotated by single 
high-level category labels that are related to the most significant content of the image. 
However, RS images typically contain multiple classes and thus each image 
can be simultaneously associated with different land-cover class labels (i.e., multi-labels). 
To overcome these problems, we introduce the BigEarthNet that is a new large-scale Sentinel-2 archive\footnote{The BigEarthNet is available at \url{http://bigearth.net}.} and contains $590,326$ Sentinel-2 image patches. Each patch is annotated with multi-labels 
provided from the CORINE Land Cover database, which is updated in 2018 (CLC 2018). 
We propose our archive as a sufficient source for RS image analysis with deep learning. 
In order to test the BigEarthNet on RS image analysis problems, we focus our attention on image scene classification. To this end, we consider a shallow CNN architecture to be trained on the BigEarthNet. We compare the results obtained by this network with the Inception-v2~\cite{Szegedy:2016} pre-trained on the ImageNet. 
We believe that it will make a significant advancement in terms of 
developments of algorithms for the analysis of large-scale RS image archives. 

\vspace{-0.12in}
\section{Limitations of Existing Remote Sensing Image Archives}
\vspace{-0.07in}
\label{sec:method}

Most of the benchmark archives in RS (UC Merced Land Use Dataset~\cite{Yang:2010}, 
WHU-RS19~\cite{Shao:2013}, RSSCN7~\cite{Zou:2015}, SIRI-WHU~\cite{Zhao:2016}, AID~\cite{Xia:2017}, NWPU-RESISC45~\cite{Cheng:2017}, RSI-CB~\cite{Li:2017}, EuroSat~\cite{Helber:2017} and 
PatternNet~\cite{Zhou:2018}) 
contain a small number of images annotated with single category labels. Table \ref{archive_list} presents the list of the existing archives. 
These archives become popular 
for the implementation, evaluation and validation of algorithms in the context of image classification, search and 
retrieval tasks. However, RS community encounters critical limitations, while using these archives for 
applying deep learning based approaches. One of the most critical limitations is that the number of annotated 
images included in the existing archives is very small. Thus, they are insufficient to train modern deep neural networks to reach a high generalization ability as the models may 
overfit dramatically when using small training sets. In details, training such networks on the 
existing archive images suffers from the problem of learning a large number of 
parameters that prevents the accurate characterization of semantic content of RS images. 
\begin{table}[t]
  \setlength{\tabcolsep}{1.5pt}
  \renewcommand{\arraystretch}{1}
  \footnotesize
  \captionsetup{justification=justified,singlelinecheck=false, format=hang}
  \caption{List of the existing RS archives.}
  \centering
  \label{archive_list}
  \begin{tabular}{lccc}
  \hline 
  \textbf{Archive Name} & \specialcell{\textbf{Image}\\\textbf{Type}} & \specialcell{\textbf{Annotation}\\\textbf{Type}} & \specialcell{\textbf{Number}\\\textbf{of Images}}\\
  \hline 
  \multirow{ 2}{*}{UC Merced}  & \multirow{ 2}{*}{Aerial RGB} & \makecell{Single Label}~\cite{Yang:2010} & 2,100\\
  & & \makecell{Multi-Label}~\cite{Chaudhuri:2018} & 2,100\\
  \hline
  WHU-RS19~\cite{Shao:2013} & Aerial RGB & Single Label & 1,005\\
  \hline
  RSSCN7~\cite{Zou:2015} & Aerial RGB & Single Label & 2,800\\
  \hline
  SIRI-WHU~\cite{Zhao:2016} & Aerial RGB & Single Label & 2,400\\
  \hline
  AID~\cite{Xia:2017} & Aerial RGB & Single Label & 10,000\\
  \hline
  \specialcellleft{NWPU-RESISC45}~\cite{Cheng:2017} & Aerial RGB & Single Label & 31,500\\
  \hline
  RSI-CB~\cite{Li:2017} & Aerial RGB & Single Label & 36,707\\
  \hline
  EuroSat~\cite{Helber:2017} & Satellite Multispectral & Single Label & 27,000\\
  \hline
  PatternNet~\cite{Zhou:2018} & Aerial RGB & Single Label & 30,400\\
  \hline
  \end{tabular}
  \vspace{-0.5cm}
\end{table}
To this end, fine-tuning the models pre-trained on ImageNet is used as a transfer learning approach. 
However, the profound differences between the image properties of computer vision and RS 
limit the accurate characterization of RS images when fine-tuning approach is applied. 
As an example, Sentinel-2 images have $13$ spectral bands associated to varying and lower spatial resolutions with respect to computer vision images. 
There are also differences in the ways that the category labels of 
computer vision and RS are defined for the semantic content of an image. Thus, 
fine-tuning pre-trained models for RS images may not be generally applicable to reduce 
this semantic gap and therefore may lead to weak discrimination ability for land-cover 
classes. Another limitation of existing archives is that they contain images annotated by 
single high-level category labels, which are related to the most significant content of
the image. However, RS images generally contain multiple classes so that they can be 
simultaneously associated to different land-cover class labels (i.e., multi-labels). Hence, 
a benchmark archive consisting of images annotated with multi-labels is required. 
Although the archive presented in~\cite{Chaudhuri:2018} contains images with multi-labels, 
the sample size of this archive is very small to be efficiently utilized for deep learning. 
Another limitation of RS image archives is that since researchers generally 
do not have free access to satellite data together with their annotation, most of the benchmark archives contain 
aerial images with only RGB image bands. Unavailability of a high number of annotated satellite images prevents to 
employ deep learning methods in a convenient way for 
the complete understanding of huge amount of freely accessible satellite data 
(e.g., Sentinel-1, Sentinel-2). Although the benchmark archive proposed in~\cite{Helber:2017} 
includes annotated satellite images, the number of images is still small. 
Aforementioned limitations of existing archives 
reveal the need for a large-scale RS benchmark archive to be used for 
training deep neural networks instead of the ImageNet.
\begin{table}[ht]
  \setlength{\tabcolsep}{13.pt}
  \renewcommand{\arraystretch}{0.87}
  \footnotesize
  \captionsetup{justification=justified,singlelinecheck=false, format=hang}
  \caption{The considered Level-3 CLC classes and the number of images associated with each land-cover class in the BigEarthNet.}
  \centering
  \label{labels}
  \begin{tabular}{lc}
  \hline 
  \textbf{Land-Cover Classes} & \makecell{\textbf{Number of}\\\textbf{Images}} \\
  \hline
  Mixed forest & $217,119$ \\ 
  \hline
  Coniferous forest & $211,703$ \\ 
  \hline
  Non-irrigated arable land & $196,695$ \\ 
  \hline
  Transitional woodland/shrub & $173,506$ \\ 
  \hline
  Broad-leaved forest & $150,944$ \\ 
  \hline
  \specialcellleft{Land principally occupied by agriculture,\\with significant areas of natural vegetation} & $147,095$ \\ 
  \hline
  Complex cultivation patterns & $107,786$ \\ 
  \hline
  Pastures & $103,554$ \\ 
  \hline
  Water bodies & $83,811$ \\ 
  \hline
  Sea and ocean & $81,612$ \\ 
  \hline
  Discontinuous urban fabric & $69,872$ \\ 
  \hline
  Agro-forestry areas & $30,674$ \\ 
  \hline
  Peatbogs & $23,207$ \\ 
  \hline
  Permanently irrigated land & $13589$ \\ 
  \hline
  Industrial or commercial units & $12895$ \\ 
  \hline
  Natural grassland & $12,835$ \\ 
  \hline
  Olive groves & $12,538$ \\ 
  \hline
  Sclerophyllous vegetation & $11,241$ \\ 
  \hline
  Continuous urban fabric & $10,784$ \\ 
  \hline
  Water courses & $10,572$ \\ 
  \hline
  Vineyards & $9,567$ \\ 
  \hline
  Annual crops associated with permanent crops & $7,022$ \\ 
  \hline
  Inland marshes & $6,236$ \\ 
  \hline
  Moors and heathland & $5,890$ \\ 
  \hline
  Sport and leisure facilities & $5,353$ \\ 
  \hline
  Fruit trees and berry plantations & $4,754$ \\ 
  \hline
  Mineral extraction sites & $4,618$ \\ 
  \hline
  Rice fields & $3,793$ \\ 
  \hline
  Road and rail networks and associated land & $3,384$ \\ 
  \hline
  Bare rock & $3,277$ \\ 
  \hline
  Green urban areas & $1,786$ \\ 
  \hline
  Beaches, dunes, sands & $1,578$ \\ 
  \hline
  Sparsely vegetated areas & $1,563$ \\ 
  \hline
  Salt marshes & $1,562$ \\ 
  \hline
  Coastal lagoons & $1,498$ \\ 
  \hline
  Construction sites & $1,174$ \\ 
  \hline
  Estuaries & $1,086$ \\ 
  \hline
  Intertidal flats & $1,003$ \\ 
  \hline
  Airports & $979$ \\ 
  \hline
  Dump sites & $959$ \\ 
  \hline
  Port areas & $509$ \\ 
  \hline
  Salines & $424$ \\ 
  \hline
  Burnt areas & $328$ \\ 
  \hline
  \hline 
  \end{tabular}
  \vspace{-0.6cm}
  \end{table}
\vspace{-0.15in}
\section{The BigEarthNet Archive}
\begin{figure}
  \captionsetup{justification=raggedright,singlelinecheck=false, format=hang, font=scriptsize}
  \begin{minipage}[c]{0.17\linewidth}
    \includegraphics[width=\columnwidth]{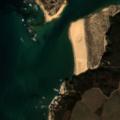}
  \end{minipage}
  \hfill
  \begin{minipage}[c]{0.31\linewidth}
    \captionsetup{justification=raggedright,singlelinecheck=false, format=hang, font=scriptsize}
    \caption*{permanently irrigated land, sclerophyllous vegetation, beaches, dunes, sands, estuaries, sea and ocean}   
  \end{minipage}
  \hfill
  \begin{minipage}[c]{0.17\linewidth}
    \includegraphics[width=\columnwidth]{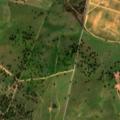}
  \end{minipage}
  \hfill
  \begin{minipage}[c]{0.31\linewidth}
    \captionsetup{justification=raggedright,singlelinecheck=false, format=hang, font=scriptsize}
    \caption*{non-irrigated arable land, fruit trees and berry plantations, agro-forestry areas, transitional woodland/shrub}
  \end{minipage}
  \hfill
  \vspace{-0.08cm}
  \begin{minipage}[c]{0.17\linewidth}
    \includegraphics[width=\columnwidth]{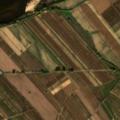}
  \end{minipage}
  \hfill
  \begin{minipage}[c]{0.31\linewidth}
    \captionsetup{justification=raggedright,singlelinecheck=false, format=hang, font=scriptsize}
    \caption*{permanently irrigated land, vineyards, beaches, dunes, sands, water courses}
  \end{minipage}
  \hfill
  \begin{minipage}[c]{0.17\linewidth}
    \includegraphics[width=\columnwidth]{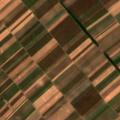}
  \end{minipage}
  \hfill
  \begin{minipage}[c]{0.31\linewidth}
    \captionsetup{justification=raggedright,singlelinecheck=false, format=hang, font=scriptsize}
    \caption*{non-irrigated arable land}
  \end{minipage}
  \hfill
  \vspace{-0.08 cm}
  \begin{minipage}[c]{0.17\linewidth}
    \includegraphics[width=\columnwidth]{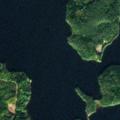}
  \end{minipage}
  \hfill
  \begin{minipage}[c]{0.31\linewidth}
    \captionsetup{justification=raggedright,singlelinecheck=false, format=hang, font=scriptsize}
    \caption*{coniferous forest, mixed forest, water bodies}
  \end{minipage}
  \hfill
  \begin{minipage}[c]{0.17\linewidth}
    \includegraphics[width=\columnwidth]{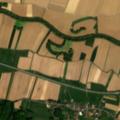}
  \end{minipage}
  \hfill
  \begin{minipage}[c]{0.31\linewidth}
    \captionsetup{justification=raggedright,singlelinecheck=false, format=hang, font=scriptsize}
    \caption*{discontinuous urban fabric, non-irrigated arable land, land principally occupied by agriculture, broad-leaved forest}
  \end{minipage}
  \hfill
  \captionsetup{justification=raggedright,singlelinecheck=false, format=hang, font={small,stretch=0.75}}
  \caption{Example of Sentinel-2 images and their multi-labels in our BigEarthNet archive.}
  \label{fig:patch_ex}
  \vspace{-0.7cm}
\end{figure}

\begin{figure}[t]
  \centering
  \captionsetup{justification=raggedright,singlelinecheck=false, format=hang}
  \includegraphics[trim=0 0 0 0,clip,width=\linewidth]{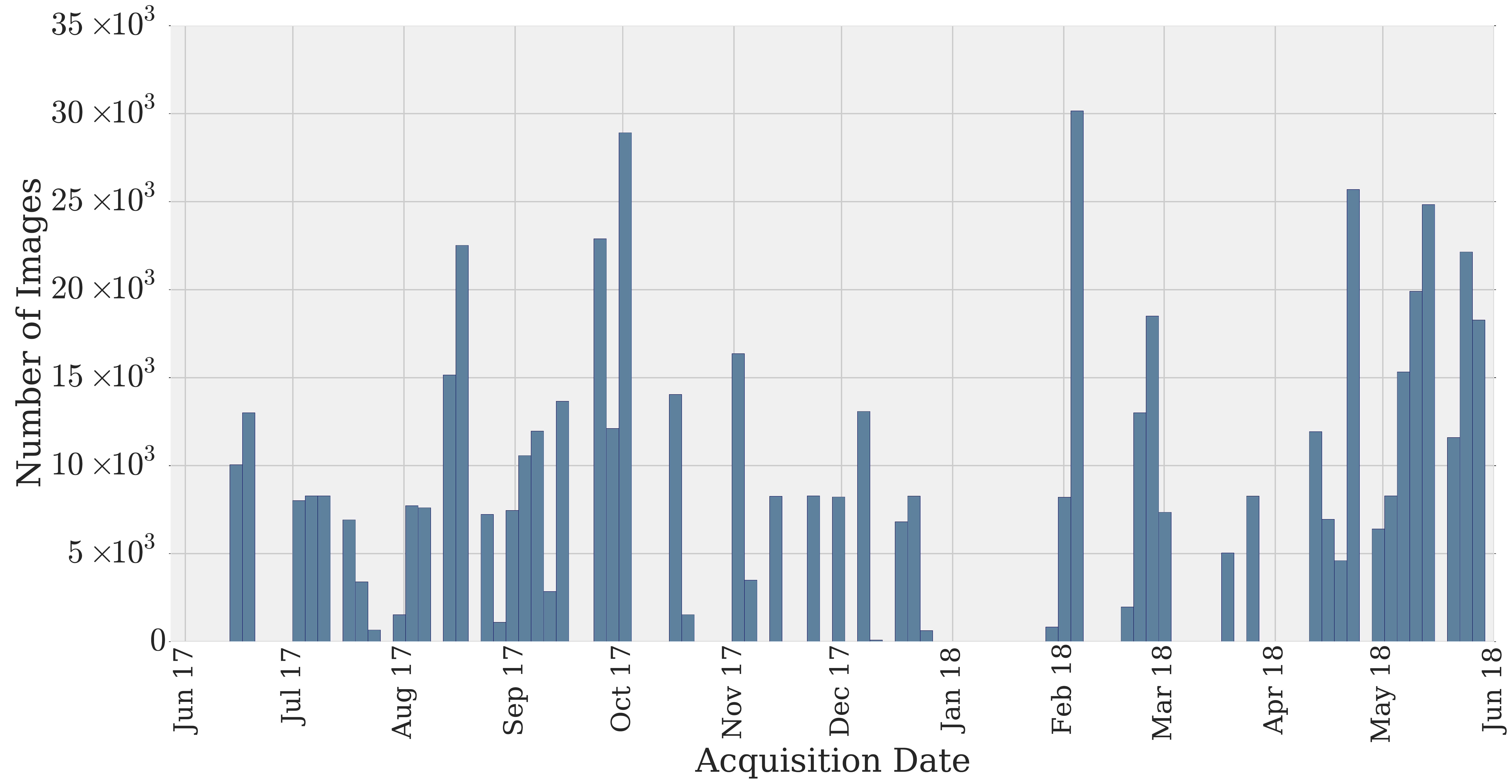} 
  \caption{The number of Sentinel-2 images with respect to acquisition date.}
  \label{fig:acquisition}
   \vspace{-0.6cm}
\end{figure}

\vspace{-0.1in}
\label{sec:exp}
To overcome the limitations of existing archives, we introduce the BigEarthNet that is the first large-scale 
benchmark archive in RS. We have constructed our archive 
by selecting $125$ Sentinel-2 tiles acquired between June 2017 and May 2018. 
Considered tiles are distributed over the $10$ countries (Austria, Belgium, Finland, Ireland, Kosovo, Lithuania, Luxembourg, 
Portugal, Serbia, Switzerland) of Europe. It is worth noting that considered tiles are 
associated to cloud cover percentage less than 1\%. All tiles were atmospherically corrected by using Sentinel-2 Level 2A product generation and 
formatting tool (sen2cor) of ESA. Among $13$ Sentinel-2 spectral bands, 10\textsuperscript{th} band, 
for which surface information is not embodied, was excluded. 
After the tile selection and preliminary processing steps were carried out, 
selected tiles were divided into $590,326$ non-overlapping image patches. 
Each patch (denoted as image hereafter) is a section of i) $120\times120$ pixels for 10m bands; ii) $60\times60$ pixels for 20m bands; and iii) $20\times20$ pixels for 60m bands. We have associated each image with 
one or more land-cover class labels (i.e., multi-labels) provided from the CORINE Land Cover (CLC) database 
of the year 2018 (CLC 2018). The CLC inventory was produced by the Eionet National Reference Centres on Land Cover 
with the coordination of the European Environment Agency (EEA) for the 
recognition, identification and assessment of land cover classes by leveraging 
the texture, pattern and density information of 
the objects presented in RS images. This inventory is very recently 
updated as CLC 2018, for which the annotation 
process has been carried out for the period of 2017-2018. We selected tiles within 
the considered time interval to be appropriate for the annotation period of CLC 2018. 
CLC nomenclature includes land cover classes grouped in a three-level hierarchy\footnote{\url{https://land.copernicus.eu/user-corner/technical-library/corine-land-cover-nomenclature-guidelines}}. The considered Level-3 CLC class labels and the number of images associated with each label are shown in Table \ref{labels}. We would like to note that the number of images for each land cover class varies significantly in the archive. The number of labels associated with each image varies between 
$1$ and $12$, whereas $95$\% of images have at most $5$ multi-labels. Only $15$ images contain 
more than $9$ labels in the BigEarthNet.    
Fig. \ref{fig:patch_ex} shows an example of images and their multi-labels, while Fig. \ref{fig:acquisition} shows the number of Sentinel-2 images with respect to the acquisition date. 
It is worth noting that we aimed to represent each considered geographic location with images acquired in all 
different seasons. However, due to the difficulties of collecting Sentinel-2 images with lower cloud 
cover percentage within a narrow time interval, it was not possible for some areas. 
The number of images acquired in autumn, winter, spring and summer seasons are $154943$, $117156$, $189276$ 
and $128951$ respectively. Since cloud cover percentage of Sentinel-2 tiles acquired in winter is generally higher 
than the other seasons, our archive contains the lowest number of images from winter season.

We also employed the visual inspection for the quality check of image multi-labels. By visual inspection, we have identified that $70,987$ images are fully covered by seasonal snow, cloud and cloud shadow\footnote{The lists of images fully covered by seasonal snow, cloud and cloud shadow are available at \url{http://bigearth.net/\#downloads}.}. We suggest not to include these images for training and test stages of the machine/deep learning algorithms, while working on scene classification, content-based image retrieval and search if only BigEarthNet Sentinel-2 images are used.
\vspace{-0.15in}
\section{Experimental Results}
\vspace{-0.1in}
\label{sec:conc}
\begin{table*}[t]
  \small	
  \setlength{\tabcolsep}{1.5pt}
  \renewcommand{\arraystretch}{0.95}
  \captionsetup{justification=justified,singlelinecheck=false, format=hang}
  \caption{Example of Sentinel-2 images with the true multi-labels and the multi-labels assigned by the Inception-v2, the S-CNN-RGB and the S-CNN-All.}
  \centering
  \label{test_result}
  \begin{tabular}{ccccc}
  \hline 
  \makecell{Test Images} & \makecell{True Multi-Label} & \makecell{Inception-v2}  & \makecell{S-CNN-RGB} & \makecell{S-CNN-All}\\
  \hline
  \begin{minipage}{0.09\linewidth}
    \vspace{0.01in}
    \includegraphics[width=\columnwidth]{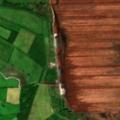}
    \vspace{-0.14in}
  \end{minipage} & 
  \begin{minipage}[c]{0.17\linewidth}
    \captionsetup{justification=centering,singlelinecheck=false, format=hang, font=small}
    \caption*{pastures, peatbogs}
  \end{minipage}
  & 
  \begin{minipage}[c]{0.22\linewidth}
    \captionsetup{justification=centering,singlelinecheck=false, format=hang, font=small}
    \caption*{non-irrigated arable land, coniferous forest, mixed forest, transitional woodland/shrub}
  \end{minipage}
  & 
  \begin{minipage}[c]{0.22\linewidth}
    \captionsetup{justification=centering,singlelinecheck=false, format=hang, font=small}
    \caption*{non-irrigated arable land, land occupied by agriculture, mixed forest}
  \end{minipage}
   & 
   \begin{minipage}[c]{0.17\linewidth}
    \captionsetup{justification=centering,singlelinecheck=false, format=hang, font=small}
    \caption*{pastures, peatbogs}
  \end{minipage}\\
  \hline
  \begin{minipage}[c]{0.09\linewidth}
    \vspace{0.01in}
    \includegraphics[width=\columnwidth]{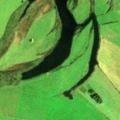}
	\vspace{-0.14in}
  \end{minipage} & 
  \begin{minipage}[c]{0.18\linewidth}
    \captionsetup{justification=centering,singlelinecheck=false, format=hang, font=small}
    \caption*{pastures, land occupied by agriculture, water bodies}
  \end{minipage} & 
  \begin{minipage}[c]{0.20\linewidth}
    \captionsetup{justification=centering,singlelinecheck=false, format=hang, font=small}
    \caption*{coniferous forest, mixed forest, transitional woodland/shrub}
  \end{minipage} &  
  \begin{minipage}[c]{0.19\linewidth}
    \captionsetup{justification=centering,singlelinecheck=false, format=hang, font=small}
    \caption*{non-irrigated arable land, land occupied by agriculture}
  \end{minipage} & 
  \begin{minipage}[c]{0.18\linewidth}
    \captionsetup{justification=centering,singlelinecheck=false, format=hang, font=small}
    \caption*{pastures, land occupied by agriculture, water bodies}
  \end{minipage} \\ 
  \hline
  \begin{minipage}[c]{0.09\linewidth}
    \vspace{0.01in}
    \includegraphics[width=\columnwidth]{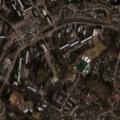}
    \vspace{-0.14in}
  \end{minipage} & 
  \begin{minipage}[c]{0.15\linewidth}
    \captionsetup{justification=centering,singlelinecheck=false, format=hang, font=small}
    \caption*{discontinuous urban fabric, industrial or commercial units}
  \end{minipage} & 
  \begin{minipage}[c]{0.18\linewidth}
    \captionsetup{justification=centering,singlelinecheck=false, format=hang, font=small}
    \caption*{coniferous forest, mixed forest, transitional woodland/shrub}
  \end{minipage} & 
  \begin{minipage}[c]{0.3\linewidth}
    \captionsetup{justification=centering,singlelinecheck=false, format=hang, font=small}
    \caption*{discontinuous urban fabric, land occupied by agriculture, broad-leaved forest, coniferous forest, mixed forest}
  \end{minipage} & 
  \begin{minipage}[c]{0.15\linewidth}
    \captionsetup{justification=centering,singlelinecheck=false, format=hang, font=small}
    \caption*{discontinuous urban fabric, industrial or commercial units}
  \end{minipage} \\ 
\hline
  \end{tabular}
  \vspace{-0.4cm}
\end{table*}
In the experiments, we have used the BigEarthNet archive in the framework of RS image scene classification 
problems. To this end, 
we selected a shallow CNN architecture, which consists of three convolutional layers with 
$32$, $32$ and $64$ filters having $5\times5$, $5\times5$ and $3\times3$ filter sizes, 
respectively. We added one fully connected (FC) layer and one classification layer to the output of last convolutional layer. In all convolution operations, zero padding was used. We also applied 
max-pooling between layers. We considered to utilize: i) only RGB channels (denoted as S-CNN-RGB); and ii) all spectral channels (denoted as S-CNN-All). For the S-CNN-All, cubic interpolation was 
applied to $20$ and $60$ meter bands of each image to have the same pixel sizes associated with each band. 
Weights of the S-CNN-RGB and the S-CNN-All were randomly initialized and we trained 
both networks from scratch on the BigEarthNet images. 
In order to show the effectiveness of the BigEarthNet to be used in training, we compared the results 
with fine-tuning one of the recent pre-trained deep learning architectures. 
\begin{table}[t]
  \setlength{\tabcolsep}{5.pt}
  \captionsetup{justification=justified,singlelinecheck=false, format=hang}
  \caption{Experimental results obtained by the Inception-v2, the S-CNN-RGB and the S-CNN-All.}
  \centering
  \label{result}
  \begin{tabular}{lcccc}
  \hline
  Method & $P$ ($\%$) & $R$ ($\%$) & $F_1$ & $F_2$\\
  \hline
  Inception-v2~\cite{Szegedy:2016} & $48.23$ & $56.79$ & $0.4988$ & $0.5301$\\
  S-CNN-RGB & $65.06$ &  $75.57$ & $0.6759$ & $0.7139$\\
  \textbf{S-CNN-All} & $\bm{69.93}$ & $\bm{77.10}$ & $\bm{0.7098}$ & $\bm{0.7384}$\\
  \hline 
  \end{tabular}
  \vspace{-0.5cm}
\end{table}
We considered the Inception-v2 network~\cite{Szegedy:2016} 
pre-trained on ImageNet as a state-of-the-art architecture. We used the feature vector 
extracted from the layer just before the softmax layer of the Inception-v2. To employ fine-tuning, 
we fixed the model weights of the Inception network. We added one FC and one 
classification layer to the network and just fine-tuned these layers by using the RGB channels of 
the BigEarthNet images. In the experiments, $70,987$ images that are fully covered by seasonal snow, cloud and cloud shadow were eliminated. Then, among the remaining images, we randomly selected: i) $60\%$ of images to derive a training set; ii) $20\%$ of images to derive a validation set; and iii) $20\%$ of images to derive a test set. Both for fine-tuning and training from scratch, we selected the number of epochs as $100$ and Stochastic Gradient Descent algorithm is employed in order to decrease the sigmoid cross entropy loss (which aims at maximizing the log-likelihood of each land-cover class throughout all training images). 
For the performance metrics of experiments, we employed precision ($P$), recall ($R$), $F1$ and $F2$ scores, 
which are widely used metrics for multi-label image classification. 
As it can be seen from Table \ref{result}, the S-CNN-RGB provides better performance than the 
Inception-v2 in all metrics, while both networks consider only RGB image channels.
When the S-CNN-All architecture is trained on the BigEarthNet images containing all spectral 
bands, the results become much more 
promising with respect to using only RGB bands. 
Table \ref{test_result} shows the example of Sentinel-2 images with the true multi-labels and 
the multi-labels assigned by the Inception-v2, the S-CNN-RGB and the S-CNN-All. 
The performance improvements on all metrics 
are statistically significant under a value of $p \ll 0.0001$. The same behavior is also observed when the BigEarthnet images are associated to Level-1 and Level-2 CLC class labels. 
We would like to also note that the S-CNN-RGB and the S-CNN-All are 
very simple CNN architectures that consist of only 3 convolutional layers and max-pooling. 
Training deeper models (which include recent 
deep learning techniques such as residual connections, wider layers with varying filter sizes etc.) from 
scratch can lead to more promising results. 
On the basis of all obtained results, we can state that 
RS community can benefit from 
these pre-trained models on the BigEarthNet instead of the computer vision archives. 
\vspace{-0.2in}
\section{Conclusion}
\vspace{-0.1in}
\label{sec:conc}
This paper presents a large-scale benchmark archive that consists
of $590,326$ Sentinel-2 image patches annotated by multi-labels 
for RS image understanding. We believe that the BigEarthNet 
will make a significant advancement for the use of deep learning in 
RS by overcoming the current limitations of the existing archives. 
Experimental results show the effectiveness of training even a simple
neural network on the BigEarthNet from scratch compared to fine-tuning 
a state-of-the-art deep learning model pre-trained on the ImageNet. 
We would like to note that we plan to regularly enrich the
BigEarthNet by increasing the number of annotated Sentinel-2 images.
\vspace{-0.2in}
\small
\section{Acknowledgements}
\vspace{-0.05in}
This work was supported by the European Research Council under the ERC Starting Grant BigEarth (759764) and the German Ministry for Education and Research as BBDC (01IS14013A).
\bibliographystyle{IEEEbib}
\small
\vspace{-0.05in}
\setstretch{0.87}
\bibliography{defs,refs}

\end{document}